\documentclass[letterpaper, 10 pt, conference]{ieeeconf}  % Comment this line out if you need a4paper

\usepackage{multirow}
\usepackage{graphicx}
\usepackage[table,xcdraw]{xcolor}
\usepackage[normalem]{ulem}
\useunder{\uline}{\ul}{}
\usepackage{url}
\usepackage{adjustbox}
\usepackage{booktabs}
\usepackage{cite}
\let\labelindent\relax
\usepackage{enumitem}
\usepackage{amsmath,amssymb,amsfonts}
\usepackage{hyperref}
\IEEEoverridecommandlockouts                              % This command is only needed if 
\title{\LARGE \bf
VIPS-Odom: Visual-Inertial Odometry Tightly-coupled with Parking Slots for Autonomous Parking
}

\author{Xuefeng Jiang$^{1,2,*}$, Fangyuan Wang$^{1,*}$, Rongzhang Zheng$^{1,*}$, Han Liu$^{1}$, \\  Yixiong Huo$^{1}$, Jinzhang Peng$^{1}$, Lu Tian$^{1}$, Emad Barsoum$^{1}$% <-this % stops a space
\thanks{*Xuefeng Jiang, Fangyuan Wang and Rongzhang Zheng share equal contributions.}% <-this % stops a space
\thanks{$^{1}$Advanced Micro Devices (AMD), Inc.
        {\tt\small \{xuefeng.jiang, fangyuan.wang, rongzhang.zheng,  han.liu, yixiong.huo, jinz.peng, lu.tian, emad.barsoum\}@amd.com}}%
\thanks{$^{2}$Institute of Computing Technology, Chinese Academy of Sciences and University of Chinese Academy of Sciences.
        {\tt\small \{jiangxuefeng21b\}@ict.ac.cn}}%
}
\begin{document}

\maketitle
\thispagestyle{plain} % 在标题页上显示页码  
\pagestyle{plain} % 在正文中显示页码 
% \thispagestyle{empty}
% \pagestyle{empty}

%%%%%%%%%%%%%%%%%%%%%%%%%%%%%%%%%%%%%%%%%%%%%%%%%%%%%%%%%%%%%%%%%%%%%%%%%%%%%%%%
\begin{abstract}
Precise localization is of great importance for autonomous parking task since it provides service for the downstream planning and control modules, which significantly affects the system performance. For parking scenarios, dynamic lighting, sparse textures, and the instability of global positioning system (GPS) signals pose challenges for most traditional localization methods. To address these difficulties, we propose VIPS-Odom, a novel semantic visual-inertial odometry framework for underground autonomous parking, which adopts tightly-coupled optimization to fuse measurements from multi-modal sensors and solves odometry. Our VIPS-Odom integrates parking slots detected from the synthesized bird-eye-view (BEV) image with traditional feature points in the frontend, and conducts tightly-coupled optimization with joint constraints introduced by measurements from the inertial measurement unit, wheel speed sensor and parking slots in the backend. We  develop a multi-object tracking framework to robustly track parking slots' states. To prove the superiority of our method, we equip an electronic vehicle with related sensors and build an experimental platform based on ROS2 system. Extensive experiments demonstrate the efficacy and advantages of our method compared with other baselines for parking scenarios.
\end{abstract}

%%%%%%%%%%%%%%%%%%%%%%%%%%%%%%%%%%%%%%%%%%%%%%%%%%%%%%%%%%%%%%%%%%%%%%%%%%%%%%%%
\section{Introduction}
\label{sec:intro}

Simultaneous localization and mapping (SLAM) is a significant task in many fields including virtual reality and autonomous driving. In the field of automotive, SLAM modules aim to achieve precise and real-time localization so that other modules like planning and control can respond on time. There are many popular visual-inertial localization estimation methods \cite{vinsmono} that obtain feature points from camera images, incorporate other sensors like the inertial measurement unit (IMU) and wheel speed sensor (WSS) to build SLAM modules, and achieve the state-of-the-art performance.
% \vspace{-2em} 
\begin{figure*}[htbp]
    \centering
    \includegraphics[width=\textwidth]{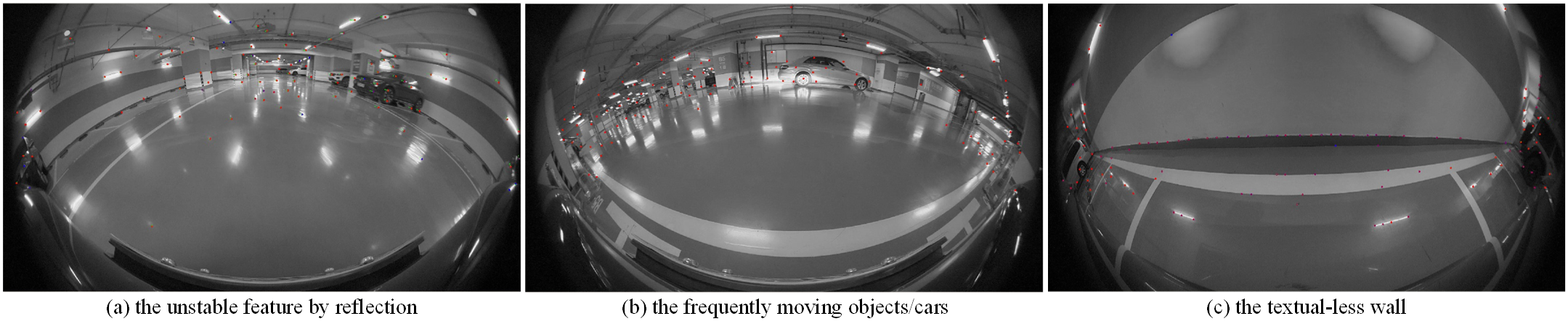}
    % \vspace{-5pt}
    \caption{Difficulties of traditional methods in typical parking environments.}
    \label{fig:difficulties}
\end{figure*}
% \vspace{-1em} 

However, as shown in Fig \ref{fig:difficulties}, traditional visual methods suffer from unstable feature tracking which is often caused by light reflection and frequently moving objects. These methods usually cannot extract enough effective feature points in texture-less and dynamic lighting parking environments, which bring challenges to autonomous parking task, including automated parking assistance (APA) and automated valet parking (AVP). APA aims to assist parking the vehicle into a nearby vacant parking slot while 
AVP focuses on long-distance navigation around the parking lot and finding a vacant parking slot.
% AVP aims to assist long-distance navigation around the parking lot and to find a vacant parking slot.
Both tasks have high requirements on localization precision.
In practice, insufficient and unstable feature points in complicated parking environments cause accumulated drifts, and even failures in extreme cases \cite{survey}. 
Though there are some methods to mitigate accumulated drifts like loop closure detection \cite{loopDetec}, it still remains a problem to design localization methods with high precision and stability in such complicated scenarios \cite{survey}.

% they are not feasible for all parking scenes, because APA mainly focuses on navigating the vehicle to a nearby vacant parking slot in a short-distance odometry while a vehicle can find a vacant PS without traversing a full loop of the parking lot in AVP tasks.

To improve localization performance for parking scenarios, a feasible strategy is to enable and fuse more sensors' measurements to accomplish more accurate and robust localization. In recent years, it has become a trend to exploit semantic elements from the scene to optimize localization estimation of the vehicle \cite{psnumbers,visslam,avp-slam}. The primary advantage of these methods is compensating for the weakness of traditional feature point detectors, and providing more constraints for the localization estimation. 
% Meanwhile, we also notice that there still lacks evaluation about how existing SLAM methods' performance for underground autonomous parking tasks, especially on the short-distance APA task. 

% related works should discuss the what is optimization-based and VI system
In this work, we propose VIPS-Odom, a semantic visual-inertial optimization-based odometry system to reduce accumulated localization errors by dynamically incorporating detected parking slot corner points as extra robust feature points into the SLAM frontend, and simultaneously exploiting parking slot observations as semantic constraints of optimization in the SLAM backend. Meanwhile, different from previous works, we robustly track and maintain all parking slots' states under a multi-object tracking framework.
Our method achieves higher localization precision against other baseline methods  for autonomous parking. Our main technical contributions can be summarized as follows:
\begin{enumerate}[leftmargin=0.4cm]
    \item To tackle difficulties in the underground environment, we propose to exploit parking slots as extra high-level semantic observation to assist SLAM. To the best of our knowledge, VIPS-Odom is the first to provide a holistic solution of fusing the parking slot observation into both the SLAM frontend and backend via an explicit way.
    \item We equip an electronic vehicle with related sensors as our experimental platform and conduct extensive real-world experiments over both long-distance and short-distance trajectories in different underground parking lots, which is more enriched than previous related works for autonomous parking scenarios.
    % We utilize multiple commercial-grade low-cost sensors to build multi-sensor SLAM method VIPS-Odom via an tightly-coupled optimization-based framework. % Notably, VIPS-Odom and other baseline methods are deployed on an electronic vehicle.
    \item Experimental results reflect the superiority and stability of VIPS-Odom against other baselines, including inertial-based EKF, traditional visual-inertial optimization-based VINS, semantic visual-inertial optimization-based VISSLAM and LiDAR-based A-LOAM.
    % Extensive experiments are conducted on multiple real-world underground parking lots.  Experimental results reflect the superior performance of VIPS-Odom against other baseline methods.  
\end{enumerate}

\section{Related Work}
\label{sec:rw}

% All text must be in a two-column format.
% The total allowable size of the text area is $6\frac78$ inches (17.46 cm) wide by $8\frac78$ inches (22.54 cm) high.
% Columns are to be $3\frac14$ inches (8.25 cm) wide, with a $\frac{5}{16}$ inch (0.8 cm) space between them.
% The main title (on the first page) should begin 1 inch (2.54 cm) from the top edge of the page.
% The second and following pages should begin 1 inch (2.54 cm) from the top edge.
% On all pages, the bottom margin should be $1\frac{1}{8}$ inches (2.86 cm) from the bottom edge of the page for $8.5 \times 11$-inch paper;
% for A4 paper, approximately $1\frac{5}{8}$ inches (4.13 cm) from the bottom edge of the
% page.
%-------------------------------------------------------------------------

\subsection{Multi-sensor SLAM}
% Simultaneous localization and mapping (SLAM) is a real-time localization and map construction technology based on sensor measurements, which is widely used in auto-drive, robotics and other fields \cite{SLAM}.
% Its core objective is to fuse multi-modal data from sensors (such as cameras, LiDAR, inertial measurement units (IMU)), and there are some well-noted methods including Extended Kalman filter (EKF) \cite{ekf}, MSCKF \cite{MSCKF}, VINS \cite{vinsmono}, ORB-SLAM \cite{ORB-SLAM} and A-LOAM \cite{loam, aloam}. 
% Among them, EKF \cite{ekf} is the classic inertial-based method to associate IMU and wheel speed sensor (WSS). A-LOAM \cite{aloam} calculates the timely odometry via LiDAR point cloud registration \cite{icp,loam}, which is an relatively expensive approach.
There are some well-noted localization methods including Extended Kalman filter (EKF) \cite{ekf}, MSCKF \cite{MSCKF}, VINS \cite{vinsmono}, ORB-SLAM \cite{ORB-SLAM} and LOAM \cite{loam, aloam}. 
EKF \cite{ekf} is a classical loosely-coupled framework to fuse measurements from various sensors. A-LOAM \cite{aloam} calculates the odometry via LiDAR point cloud registration \cite{loam} which requires the expensive LiDAR sensor.
The high cost of LiDAR sensors makes it an impractical choice for standard L2 parking applications.
For commercial-grade autonomous parking systems, it is preferred to use the camera, inertial measurement unit (IMU) and wheel speed sensor (WSS) instead of the expensive LiDAR. Thus, visual-inertial SLAM (VI-SLAM) is considered to be more promising  for commercial use. 
According to different approaches to utilizing multi-sensor measurements, VI-SLAM systems can be categorized as loosely-coupled and tightly-coupled methods according to their ways to fuse measurements from different sensors. 
The former updates the pose estimation with measurements from various sensors independently, while the latter tightly associates those measurements to jointly optimize the odometry, which often leads to more precise localization \cite{visslam}. 
% One representative state-of-the-art and open-source method is VINS
VINS \cite{vinsmono} is a representative state-of-the-art and open-source method, which utilizes the front-view camera and IMU to jointly optimize odometry in a sliding window composed of several keyframes. 
VI-SLAM system depends on low-level visual features points or directly utilize the intensity of pixels. However, both low-level visual features and image pixels suffer from instability and inconsistency during navigating in parking environments \cite{avp-slam,visslam}.  

%-------------------------------------------------------------------------
\subsection{Semantic SLAM for Autonomous Parking}
% Underground parking lots are usually decorated with enriched land markings (like parking slot lines \cite{visslam}, parking slot numbers \cite{psnumbers} and bumper lines \cite{avp-slam}), which can be utilized as semantic features to improve odometry precision. 
% Semantic-incorporated SLAM is crucial for autonomous driving, particularly autonomous parking applications. 
% There are some attempts to incorporate semantic features into VI-SLAM systems \cite{slam++,recovering,mask-slam}. SLAM++ \cite{slam++} is an early attempt to build a real-time object-oriented SLAM system which constructs an explicit graph of objects. \cite{recovering} proposes a SLAM system where semantic objects in the environment are incorporated into a bundle-adjustment framework. Mask-SLAM \cite{mask-slam} improves feature point selection with the assistance of semantic segmentation to exclude unstable objects.
% However, the methods mentioned above do not consider the complicated scenario in the underground parking lot.
There are some attempts to incorporate semantic features into VI-SLAM systems. SLAM++ \cite{slam++} is an early attempt to build a real-time object-oriented SLAM system which constructs an explicit graph of objects. \cite{recovering} proposes a SLAM system where semantic objects in the environment are incorporated into a bundle-adjustment framework. Mask-SLAM \cite{mask-slam} improves feature point selection with the assistance of semantic segmentation to exclude unstable objects. These methods usually consider semantic features from traditional objects to assist the SLAM task. However, they may be not suitable in the complicated parking environments since parking lots are decorated with scene-specific unconventional semantic markings (like parking slot lines \cite{visslam}, parking slot numbers \cite{psnumbers} and bumper lines \cite{avp-slam}).

For autonomous parking scenarios, AVP-SLAM \cite{avp-slam} is a representative semantic SLAM system, which includes the mapping procedure and the localization procedure. In the mapping process, it constructs a semantic map including parking lines and other semantic elements segmented by a modified U-Net \cite{unet}. Based on this constructed map and inertial sensors, poses of the vehicle can be estimated in the localization procedure via a loosely-coupled way. However, the localization precision would drop if the constructed map is not frequently updated since the appearance of the parking lot will change as time goes by.

Different from AVP-SLAM, there are some attempts to utilize tightly-coupled approaches by exploiting semantic objects. 
% match policy of VISSLAM
VISSLAM \cite{visslam} exploits parking slots as semantic features in the SLAM backend optimization for the first time. However, its system utilizes DeepPS from Matlab to detect parking slots (PSs) and resorts to a hard-match based strategy to match the previously observed PSs, which makes it prone to mismatching . To build a more robust SLAM system for parking and overcome the limitations of VISSLAM, we build our VIPS-Odom system and utilize the ubiquitous YOLOv5 \cite{yolov5} to perform real-time PS detection, flexibly and robustly track and maintain all PS states via a multi-object tracking framework instead of the hard-match strategy. To overcome difficulties of parking environments, we further adopt the observed PSs as robust semantic constraints in the SLAM backend and add those robust feature points to the frontend to achieve more stable and robust localization. We additionally add wheel encoder measurements into backend optimization to further improve the localization accuracy. We conduct more diverse comparison than two most relevant works AVP-SLAM and VISSLAM as shown in Tab.\ref{tab:eval}.
\vspace{-1em}  
\begin{table}[htbp]
\caption{Evaluation diversity of our study.}
\label{tab:eval}
\begin{adjustbox}{width=\columnwidth,center}
\begin{tabular}{c|c}
\toprule \midrule
Methods & Compared Baseline Methods for Localization Evaluation\\ \hline
AVP-SLAM \cite{avp-slam} & ORB-SLAM2 (Camera) \\ \hline
VIS-SLAM \cite{visslam}& - \\ \hline
Ours &  EKF (IMU+WSS) , A-LOAM (LiDAR) , VINS (IMU+Camera+WSS), VIS-SLAM (IMU+Cameras)  \\ \hline
\end{tabular}
\end{adjustbox}
\end{table}
\vspace{-1em}  

\section{Methodology}
\label{sec:methodology}

% All text must be in a two-column format.
% The total allowable size of the text area is $6\frac78$ inches (17.46 cm) wide by $8\frac78$ inches (22.54 cm) high.
% Columns are to be $3\frac14$ inches (8.25 cm) wide, with a $\frac{5}{16}$ inch (0.8 cm) space between them.
% The main title (on the first page) should begin 1 inch (2.54 cm) from the top edge of the page.
% The second and following pages should begin 1 inch (2.54 cm) from the top edge.
% On all pages, the bottom margin should be $1\frac{1}{8}$ inches (2.86 cm) from the bottom edge of the page for $8.5 \times 11$-inch paper;
% for A4 paper, approximately $1\frac{5}{8}$ inches (4.13 cm) from the bottom edge of the
% page.
%-------------------------------------------------------------------------
\subsection{Sensor Equipment \& Calibration}
\label{sec:sensors}
% \subsection{VIPS-Odom}
% Multi-modal sensor data of VIPS-Odom include visual feature points detected from front-view fisheye camera, pre-integration \cite{vinsmono} of IMU and WSS measurements between two consecutive keyframes, and parking slot (PS) observations from BEV images. 
Multi-modal sensor data of VIPS-Odom include visual feature points derived from corner detection, IMU and WSS measurements between two consecutive keyframes \cite{vinsmono}, and parking slot (PS) observations from BEV images. 
% High-quality calibration of these sensors is the key to accurate pose estimation.
% This joint optimization model acts as a critical role in fusing multi-modal sensor information to conduct backend optimization.

\begin{figure}[t]
    \centering
    \includegraphics[width=\linewidth]{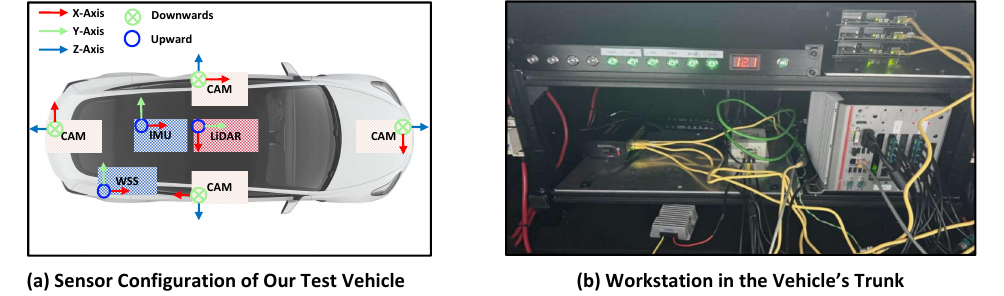}
    \caption{A brief illustration of our sensor configuration.}
    \label{fig:configure}
\end{figure}

\textbf{Sensor Configuration} Our system (in Fig.\ref{fig:configure}) includes an inertial measurement unit (IMU, mounted in the center of this vehicle), a wheel speed sensor (WSS, mounted near this vehicle's right back wheel), and four surround-view fisheye cameras mounted at the front, rear, left, and right side. Images are recorded with a resolution of $1920 \times 1080$ pixels. The BEV images with resolution of $576 \times 576$ pixels cover an area of $11.32m \times 11.32m$. 
% A $360^\circ$ LiDAR is also equipped to support evaluation of A-LOAM \cite{loam,aloam}. 
A $360^\circ$ LiDAR is also mounted on the vehicle roof which produces point cloud data to support LiDAR-SLAM evaluation for A-LOAM \cite{aloam}. 
% support evaluation of A-LOAM \cite{loam,aloam}. 
% Its details will be thoroughly presented as follows with regard to its formulation and all loss factors during optimization.

\textbf{Sensor Calibration}
Since various sensor measurements are tightly-coupled in VIPS-Odom, we carefully calibrate related parameters. 
We utilize an open-source toolbox \cite{toolbox} to optimize intrinsic of four fisheye cameras. For extrinsic calibration, we follow \cite{extrinsics} to refine the camera-IMU and IMU-WSS transformation with coarse initialization.
% Sensor calibrations can be categorized into three aspects, surround-view fisheye cameras, camera-IMU, and IMU-WSS calibration. Different from VINS \cite{vinsmono}, we conduct calibration in an offline manner, wherein calibrations are done before running VIPS-Odom system.
% We utilize an open-source toolbox \cite{toolbox, DBLP:conf/iros/HengLP13} to conduct calibration of fisheye cameras' intrinsics, and follow \cite{extrinsics} to conduct extrinsic calibrations.
\vspace{-1em} 

% Do we have to plot this?
% There are four types of coordinate systems (or frames) in our methods including BEV image frame, front-view fisheye image frame, vehicle's body frame and vehicle's world frame illustrated in figure X.
% \subsection{Synthesizing BEV Images}
\subsection{BEV Image Generation}
\label{sec:bev}
We adopt inverse perspective mapping (IPM) \cite{avp-slam} to construct the bird's eye view (BEV) image with four surround-view fisheye cameras. Relationships of the body coordinate, image coordinate and BEV coordinate can be formulated as:

\vspace{-1em} 
\begin{equation}
\left[\begin{array}{c}
            x_{v} \\
            y_{v} \\
            1
          \end{array}
    \right]
=T_{c}^{v}\pi^{-1}\left(\left[\begin{array}{c}
            u_{f} \\
            v_{f} \\
            1
          \end{array}
    \right]\right)
=H_{b}^{v}\left[\begin{array}{c}
            u_{b} \\
            v_{b} \\
            1
          \end{array}
    \right]
\label{eq:bev_map0}
\end{equation}

where $\pi^{-1}(\cdot)$ is the inverse of fisheye projection model. $T_{c}^{v}$ is the extrinsic matrix of camera with respect to the body coordinate, and $H_{b}^{v}$ denote the transformation from the BEV coordinate to body coordinate. $[x_{v},y_{v}]^{\top}$, $[x_{f},y_{f}]^{\top}$ and $[x_{b},y_{b}]^{\top}$ stand for locations at the body, fisheye camera and BEV coordinate respectively.
\subsection{Optimization Formulation}
\label{sec:optimization}
\begin{figure*}[t]
    \centering
    \includegraphics[width=\textwidth]{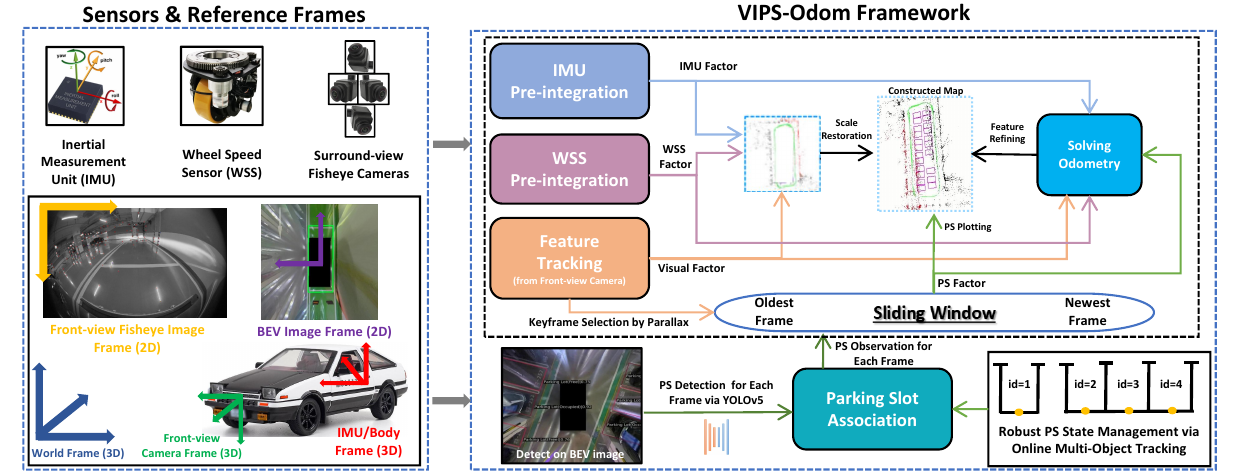}
    % \caption{Framework of VIPS-Odom. By aligning pre-integrated IMU and WSS measurement with the visual features in the front-view image, a map with metric scale can be obtained. Parking slot observation serves as semantic information in the constructed map.}
    \caption{Framework of VIPS-Odom. Our system integrates parking slot observation with measurements from IMU/WSS together for tightly-coupled optimization. By introducing constraint from parking slot observation, the system achieves higher accuracy and stability.}
    \label{fig:methodd}
\end{figure*}
% \vspace{-1.2em} 

Generally, VIPS-Odom is a tightly-coupled optimization-based SLAM system which integrates measurements from multi-modal sensors. The overall framework of VIPS-Odom is shown in Fig.\ref{fig:methodd}.
Herein we illustrate the optimization framework of VIPS-Odom. $\mathcal{Z}$ denotes the visual feature points captured in front-view fisheye camera, $\mathcal{O}$ denotes the observation of parking slot center in the world frame, $\mathcal{M}$ and $\mathcal{W}$ denote the IMU measurements and WSS measurements. We aim to solve the poses $\mathcal{T}$, map keypoints $\mathcal{P}$ matched with $\mathcal{Z}$ and the locations of PS center $\mathcal{L}$. The optimization objective can be defined as follows:
\vspace{-0.5em}
\begin{equation}
\small
\{\mathcal{L},\mathcal{T},\mathcal{P}\}^{*}=\arg\max_{\mathcal{L},\mathcal{T},\mathcal{P}}p(\mathcal{L},\mathcal{T},\mathcal{P}|\mathcal{O},\mathcal{Z},\mathcal{M},\mathcal{W}).
\end{equation}
Our main target is to get more precise estimation of $\mathcal{T}$ via multi-sensor observation. Following \cite{visslam}, we reformulate $p$ with Bayes' theorem \cite{bayesian} as follows:
% \small
\vspace{-0.5em}
\begin{equation}
\small
p(\mathcal{L},\mathcal{T},\mathcal{P}|O,\mathcal{Z},\mathcal{M},\mathcal{W})
% &=\frac{p(\mathcal{L},\mathcal{T},\mathcal{P})p(O,\mathcal{Z},\mathcal{M},\mathcal{W}|\mathcal{L},\mathcal{T},\mathcal{P})}{p(O,\mathcal{Z},\mathcal{M},\mathcal{W})} \nonumber \\
% &
\propto p(\mathcal{L},\mathcal{T},\mathcal{P})p(O,\mathcal{Z},\mathcal{M},\mathcal{W}|\mathcal{L},\mathcal{T},\mathcal{P}). 
\end{equation} 
% \end{align}
% \begin{align}
%     P(Y|do(X))&=\sum_{c\in A} P(Y|X,M)P(c) \nonumber \\
%     & = \sum_{c\in A} P(Y|X,E(X,c))P(c)
% \end{align}\label{eq:2}
We further factorize $p$ by separating $\mathcal{O}$ from other modalities since visual keypoints $\mathcal{Z}$ and PS observations $\mathcal{O}$ are obtained from two independent sensor measurements:
\vspace{-1em} 
\begin{align}
&p(\mathcal{L},\mathcal{T},\mathcal{P}|O,\mathcal{Z},\mathcal{M},\mathcal{W}) \nonumber \\
&\propto p(\mathcal{L})p(\mathcal{T},\mathcal{P})p(O|\mathcal{L},\mathcal{T},\mathcal{P})p(\mathcal{Z},\mathcal{M},\mathcal{W}|\mathcal{L},\mathcal{T},\mathcal{P}) \nonumber \\
&=p(\mathcal{L})p(\mathcal{T},\mathcal{P})p(O|\mathcal{L},\mathcal{T})p(\mathcal{Z},\mathcal{M},\mathcal{W}|\mathcal{T},\mathcal{P}) \\
% &=p(\mathcal{T},\mathcal{P})p(\mathcal{Z},\mathcal{M}|\mathcal{T},\mathcal{P})p(\mathcal{L})p(\mathcal{O}|\mathcal{L},\mathcal{T}), \nonumber \\
&=\underbrace{p(\mathcal{T})p(\mathcal{P})p(\mathcal{Z}|\mathcal{T},\mathcal{P})p(\mathcal{M},\mathcal{W}|\mathcal{T})}_{\text {visual-inertial term}}
\underbrace{p(\mathcal{L})
p(O|\mathcal{L},\mathcal{T})}_{\text {PS term}}, \nonumber
% &=\underbrace{p(\mathcal{T},\mathcal{P})p(\mathcal{Z},\mathcal{M}, \mathcal{W}|\mathcal{T},\mathcal{P})}_{\text {visual-inertial term}} \underbrace{p(\mathcal{L})p(O|\mathcal{L},\mathcal{T})}_{\text{PS term}}, 
\end{align}
where the first bracket is related to visual features, IMU/WSS motion data, and the remainder denotes the PS related term. 
% Following \cite{mapreuse,visslam}, the visual-inertial term can be converted into a visual error term and inertial error terms, $E_V$ ,$E_I$ and $E_W$, respectively. $E_V$ links each keypoint and its projecting map point while $E_I$ and $E_W$ constrain consecutive keyframes by visual-inertial alignment, predicting reliable camera pose estimation and map point locations using IMU and WSS. 
% PS observations in consequent keyframes in BEV images encode semantic information. Since PS with same id is observed in multiple consequent keyframes (i.e. the PS observation with same id observed in multiple keyframes should be very close in the world frame), it is easy to derive a registration constraint $E_{PS}$.  
% Therefore, to conduct odometry estimation, we jointly optimize visual, inertial and PS terms in a tightly-coupled objective:
% \begin{equation}
% \{\mathcal{L},\mathcal{T},\mathcal{P}\}^{*}=\arg\min_{\mathcal{L},\mathcal{T},\mathcal{P}}\mathbf{E}_{V}+\mathbf{E}_{I}+\mathbf{E}_{W}+\mathbf{E}_{PS}. \label{eq:obj}
% \end{equation}
Given above analysis, we can perform optimization by jointly minimizing visual reprojection error, inertial motion error, and PS error. VIPS-Odom deals with both low-level geometric/motion data as well as higher-level semantic information simultaneously, which facilitates more stable and accurate localization for parking scenarios. The backend optimization is discussed in Sec.\ref{sec:backend}.

\subsection{Parking Slot (PS) Detection \& Management}
\label{psdet}
\textbf{PS Detection} Different from feature line detection in AVP-SLAM \cite{avp-slam} and matlab-based model in VISSLAM \cite{visslam}, we exploit CNN for PS detection since there are previous successful attempts of using CNN to detect PS \cite{DeepPS,expert}. Following \cite{expert}, we develop our light-weight PS detector based on YOLOv5 \cite{yolov5} to detect PS in BEV images and predict the state of PS (i.e., occupied or vacant, see Fig.\ref{fig:methodd}). Detected PS states later assist the autonomous parking planning to choose a vacant PS. Training details are introduced in Sec.\ref{sec:exp_setup}.

% Since our detector only detects the entering points of one parking slot, we complete all corner points of this parking slot based on the prior knowledge of the width and height of parking slots, and kindly note that different parking lots equip PS with different sizes. 
% \subsection{Parking Slot Association \& Tracking}
% \label{method:aNT}
\textbf{PS Management}
We utilize a classic multi-object tracking framework SORT \cite{sort} to manage PS states. Originally, parking slots are detected in the BEV image frame. Then detected PS corner points are converted to the body frame of ego vehicle (see Eq.\ref{eq:bev_map0}).
According to the timestamp when the PS is detected in the BEV images, we can fetch back the nearest pose of this vehicle among the keyframes in the sliding window (see Sec.\ref{sec:backend}). Then the pose of PS could be converted from the body frame to the world frame.
The world frame is initialized at the beginning of the trajectory (see Sec.\ref{sec:sensors}).
Each parking slot can be viewed as different rectangles located in different positions in the world frame.
They are also plotted on the constructed map and BEV images (see Fig.\ref{fig:rviz}), which assist downstream  tasks in autonomous parking. 
% In assigning detections to existing parking slots, 
To assign detections to existing parking slots, we need to update the tracker status timely.
Each target's bounding box geometry is estimated by predicting its new location in the world frame based on Kalman Filter \cite{KF}. 
The assignment cost matrix is then computed as the intersection-over-union (IoU) distance between each detection and all predicted bounding boxes from existing targets. The assignment is solved with the Hungarian assignment algorithm \cite{hungarian,sort}. Additionally, a minimum IoU is imposed to reject assignments where the detection to target overlap is less than $0.3$. Matched PS detection to existing PS will share the same id. We also implement and evaluate hard match association method used in VIS-SLAM \cite{visslam}, which pre-defines thresholds for previous PS association and new PS creation, and related analysis is discussed in Sec.\ref{sec:visslam}.

\subsection{Frontend Integration with PS}
For frontend visual odometry, visual features are detected via Shi-Tomasi algorithm \cite{shitomasi} and we mildly maintain about one hundred feature points for each new image. Visual features are tracked by the KLT sparse optical flow \cite{KLT}. 
Note that feature point number is much smaller than other optimization-based SLAM methods because parking lots are often texture-less and dynamic lighting \cite{avp-slam} which implies fewer feature points that can be detected.
Correspondingly, traditional SLAM methods like ORB-SLAM \cite{ORB-SLAM} usually show unsatisfying performance in parking scenarios \cite{challenge}.

\begin{figure}[t]
    \centering
    \includegraphics[width=\linewidth]{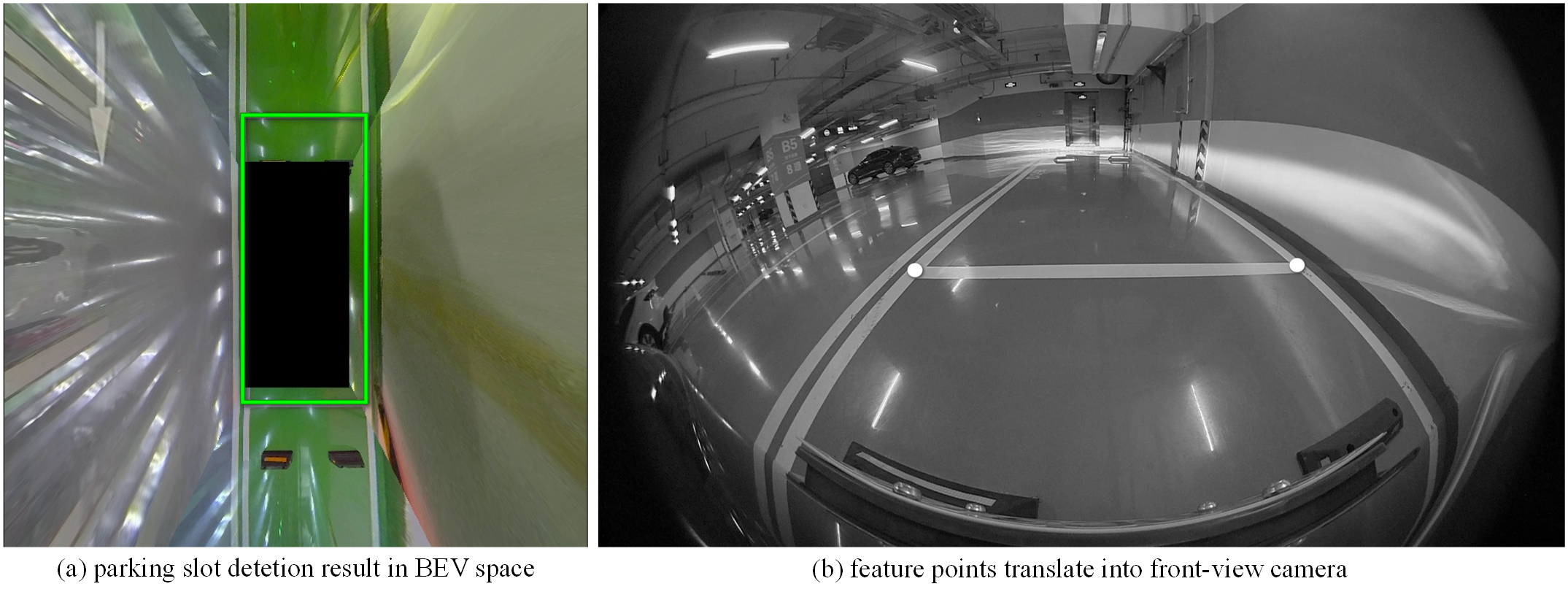}
    \caption{Feature points acquisition of frontend. (a) is parking slot detection results on the BEV image synthesized via IPM. (b) is parking slot feature points in the front-view fisheye, which is obtained by inverse projection.}
    \label{fig:front}
\end{figure}

As shown in Fig.\ref{fig:front}, we introduce detected parking slot (PS) corner points as extra robust feature points to mitigate the aforementioned issue.
We integrate fisheye camera images from multiple perspectives into one BEV image through IPM (Sec. \ref{sec:bev}). Here, the points on each BEV image correspond one-to-one to points on the original fisheye camera, therefore, each PS feature points in BEV space is paired with a pixel in front-view fish-eye camera through transformation matrix.
After predicted by the PS detector (see Sec.\ref{psdet}), these PS feature points in the BEV image frame can be inversely projected into the camera frame as extra feature points according to the transformation matrix between the BEV coordinate and the fisheye camera coordinate(see Sec.\ref{sec:bev}).
Aiming to maintain higher-quality and more robust feature points, we merge these extra PS feature points to the original detected Shi-tomasi visual feature point sequence dynamically.
These extra feature points benefit localization under the challenging texture-less environment of parking scene.
% The optical flow window for feature tracking is fixed as $50\times50$.  
2D features are firstly undistorted and then projected to a unit sphere after passing outlier rejection following VINS \cite{vinsmono}.
Keyframe selection is conducted during the SLAM frontend and we follow two commonly-used criteria in VINS \cite{vinsmono}, mainly according to the parallax between current frame and previous frames. We also conduct related experiments to examine 
% how the selection of feature point number affects our system’s precision in Sec.\ref{:comparison_robust}. 
how the number of feature point affects our system’s precision in Sec.\ref{sec:comparison_robust}. 

\subsection{Backend Optimization with PS}
\label{sec:backend}
We elaborate on the backend non-linear optimization to exploit multi-modal sensors' measurements. Based on VINS \cite{vinsmono}, VIPS-Odom performs optimization within a sliding window. Overall optimization objective can be roughly divided into visual-inertial term and PS term (see Sec.\ref{sec:optimization}). 

\textbf{Visual-inertial Term}
Visual-inertial term can be constructed via the formulation of VINS, and we further exploit the wheel speed sensor (WSS) to enrich inertial measurements for complicated environments. Specifically, in each sliding window, VIPS-Odom maintains a set of $K$ keyframes $f_0,\ldots,f_{K-1}$ to perform backend optimization. Each keyframe $f_k$ is associated with related timestamp $t_k$ of frame $k$ and camera pose $\mathcal{T}^{k} = [R_{k},t_{k}]\in\mathbb{SE}(3)$.
% , i.e. the position transformation of the keyframe $k$ in the world frame, composed of a 3×3 rotation matrix $R_{k}\in\mathbb{S}\mathbb{O}(3)$ and a 3D translation vector $t_{k}\in\mathbb{R}^3$, following the Lie algebra \cite{lie} utilized in previous works  \cite{vinsmono,cnn-slam}. 
The motion (related to change of orientation, position and velocity) between two consecutive keyframes can be determined by either pre-integrated IMU/WSS data or visual odometry \cite{vinsmono}. Referring to \cite{vinsWss}, we synthesize measurements of IMU and WSS by pre-integrating. The overall visual-intertial optimization function is formulated as follows:
\vspace{-0.5em}
\begin{equation}
\small
\mathcal{F}(\mathbf{x})=\sum_p\sum_j\mathbf{e}_r^{p,j^T}\mathbf{W}_r\mathbf{e}_r^{p,j}+\sum_{k=0}^{K-2}\mathbf{e}_s^{k^T}\mathbf{\Sigma}_{k,k+1}^{-1}\mathbf{e}_s^k+\mathbf{e}_m^T\mathbf{e}_m, \label{eq:vi_cost}
\end{equation}
where $\mathcal{F}(\mathbf{x})$ denotes the visual-inertial backend loss function, $p$ represents the index of observed feature points (or landmarks), and 
% 修改表达
$j$ is the index of the image
% belongs to the set of images 
on which the landmark $p$ appears. $K$ is the number of images in the sliding window. $e_{r}^{p,j}$ means the bundle adjustment \cite{ba} reprojection residual, $e_s^k$ represents the inertial residual, and $e_m$ denotes the marginalization residual. $\mathbf{W}_r$ is the uniform information matrix for all reprojection terms and $\boldsymbol{\Sigma}_{k,k+1}$ denotes the forward propagation of covariance \cite{vinsmono,vinsWss}. The vector $\mathbf{x}$ incorporates the motion states of keyframes, each landmark's inverse depths \cite{vinsmono} 
which can be formulated as:
\vspace{-0.5em}
\begin{equation}
    \mathbf{x}=\begin{bmatrix}\mathbf{x}_0,\mathbf{x}_1,\ldots\mathbf{x}_{K-1},\lambda_0,\lambda_1,\ldots\lambda_{m-1}\\\end{bmatrix}, \label{eq:state_vec}
\end{equation}
\begin{equation}
\mathbf{x}_k=\begin{bmatrix}\mathbf{p}_{k}^w,\mathbf{v}_{k}^w,\mathbf{q}_{k}^w,\mathbf{b}_{a_k},\mathbf{b}_{\boldsymbol{\omega}_k}\end{bmatrix}, \label{eq:state_each}
\end{equation}
% \vspace{1em}
where the state of $k$-th keyframe is $\mathbf{x}_k$, $\lambda_0,\ldots,\lambda_{m-1}$ denote the inverse depth of each landmark in camera frame,$\mathbf{p}_{k}^w,\mathbf{v}_{k}^w,\mathbf{q}_{k}^w$ denote the pose, speed and orientation of the $k$-th frame which correlates to $\mathcal{T}^k$, $\mathbf{b}_{a_k},\mathbf{b}_{\boldsymbol{\omega}_k}$ denote the IMU's accelerometer and gyroscope bias corresponding to the $k$-th image. 
% wherein technical details to conduct optimization are explained in VINS \cite{vinsmono}.
Other details of our optimization procedure are basically in accordance with  VINS \cite{vinsmono}.
 
\begin{figure}[t]
    \centering
    \includegraphics[width=\linewidth]{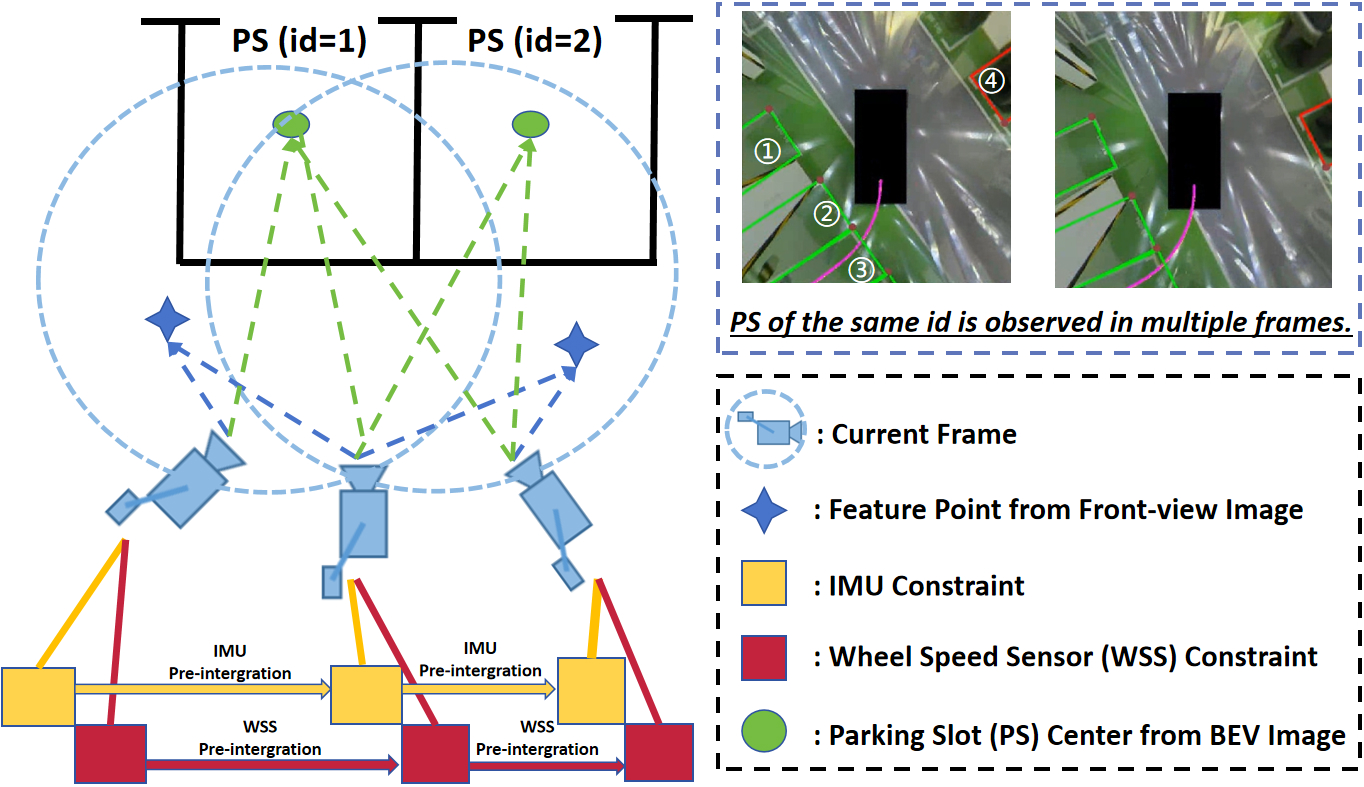}
    \caption{An illustration of our backend optimization. Note that in the BEV frame, the center point of parking slot \#2 is closer to the BEV center point than other slots (i.e. \#1,\#3, and \#4).}
    \label{fig:opt}
\end{figure}

\textbf{PS Term}
We incorporate PS into backend explicitly. In each sliding window, since each PS with the same id is associated with multiple observations, it naturally forms a registration constraint between each observed PS and its maintained state in the world frame. Each maintained state and PS observations are associated via PS id provided by our PS management module (discussed in Sec.\ref{psdet}). Our PS term is illustrated in Fig.\ref{fig:opt}. 
% The PS factor is utilized in each sliding window as semantic constraint. 
Thus, the registration error of the parking slot $i$ observed at keyframe $k$ can be defined as:
\vspace{-1em}
\begin{equation}
\mathbf{e_{reg}}^{k,i}=\mathcal{T}^k\mathcal{L}^{i,k}-\mathcal{O}_i,
\end{equation}
% \vspace{-0.5em}
where $\mathcal{O}_i$ denotes the current maintained PS (id=$i$) location state in the world frame, $\mathcal{T}^k$ denotes the $k$-th keyframe's vehicle's pose (calculated by position and orientation in Eq.\ref{eq:state_each}) in the world frame, $\mathcal{L}^{i,k}$ denotes the PS (id=$i$) observation in the body frame. PS error term is derived as:
\vspace{-0.5em}
\begin{equation}
\mathcal{G}(\mathbf{x})=\sum\limits_{k=1}^{K}\sum\limits_{i\in \mathcal{S}}^{|\mathcal{S}|}\rho((\mathbf{e_{reg}}^{k,i})^{-1}\alpha_{k}^{i}\mathbf{e_{reg}}^{k,i}),
\end{equation}\label{eq:reg}
where $S$ is the set of seen PSs in this sliding window and $\rho(\cdot)$ denotes the robust kernel function \cite{ceres}. $\alpha_k^i$ is a reweighting term based on the $k-th$ keyframe's observation of a certain parking slot (id=$i$). If the $i-th$ PS in the $k-th$ keyframe is not observed, $\alpha_k^i$ will be $0$. The reason to add this reweighting term is that due to perspective distortion and lens distortions, outer regions of generated BEV images are often stretched or compressed, which introduces more errors to PS detection (see Fig.\ref{fig:opt}). For a certain $k-th$ keyframe, we first compute the pixel distance $d_{k}^{i}$  (divided by half of BEV image size in Sec.\ref{sec:sensors}) between the center point of observed $i-th$ PS in this keyframe (supposing $k-th$ keyframe incorporates $N_k$ PSs) and the center point of BEV image in BEV image frame:
% \vspace{-1em}
\begin{equation}
\alpha_k^i = \frac{N_{k}e^{-d_{i}^{k}}}{\sum_{i=1}^{N_k} e^{-d_{i}^{k}}}.
\end{equation}
% \vspace{-1em}
Given above, $\mathcal{F}(\mathbf{x})$ covers the visual-intertial constraint term while $\mathcal{G}(\mathbf{x})$ provides a semantic registration constraint term, and their combination follows the formulation in Sec.\ref{sec:optimization}.

\section{Experiments}
\label{sec:exp}
\subsection{Experiment Setup}
\label{sec:exp_setup}
\begin{figure}[htbp]
    \centering
    \includegraphics[width=\linewidth]{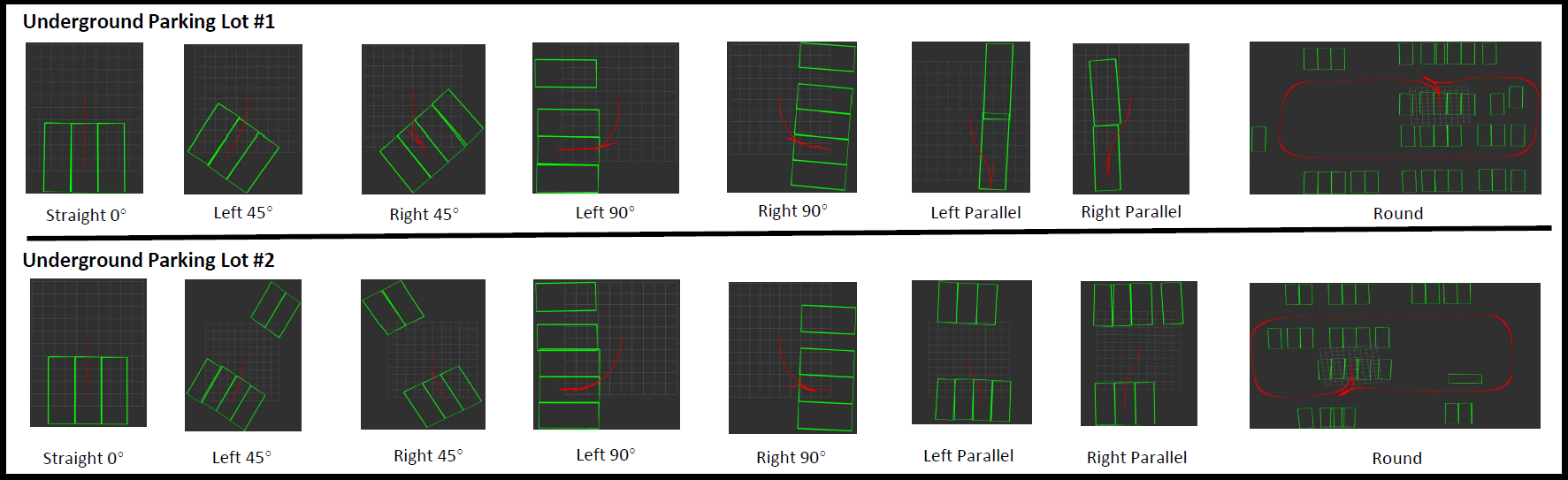}
    \caption{A visualization of collected trajectories. Rectangle regions denote parking slots captured by our system.}
    \label{fig:traj}
\end{figure}

% We conduct experiments for evaluating the proposed VIPS-Odom and other baselines.
We deploy our system on an electronic vehicle along with a workstation of Ubuntu operating system, and all modules communicate via ROS2 \cite{ROS}. The CPU and GPU configurations are i7-8700 CPU@3.20GHz and A5000 respectively. The overall optimization backend is implemented via Ceres Solver \cite{ceres} and we adopt commonly-used Cauchy function as robust kernel function (see Eq.\ref{eq:reg}). All experiments are repeated for three times and averaged. Our experiments are conducted in two parking lots (see Fig.\ref{fig:traj}). 

\textbf{Baselines}
Our baselines include four classical methods which are suitable to deploy on a commercial-grade vehicle: (i) EKF \cite{ekf}: EKF is a classic loosely-coupled approach to fusing IMU and WSS meaurements. (ii) A-LOAM \cite{aloam}: A-LOAM is a LiDAR-based method which utilizes ICP \cite{loam} with a point-to-edge distance. (iii) VISSLAM \cite{visslam}: VISSLAM utilizes IMU, feature points and matched parking slots to jointly optimize the odometry. Since its source codes are not released, we refer to VINS \cite{vinsmono} to process IMU and camera measurements, and additionally add the PS measurements into the backend optimization. (iv) VINS \cite{vinsmono}: VINS provides a tightly-coupled optimization-based approach to fusing IMU measurements and feature points from the front-view fisheye camera. In addition, we refer to \cite{vinsWss} to add WSS measurements into VINS to improve the accuracy and stability of localization for parking environments.
We compare these methods in Tab.\ref{tab:comparison}. Geometric Map is constructed by feature points, Point Cloud Map is constructed by LiDAR point clouds, and Semantic Map is constructed by both feature points and parking slots in our method.
% \vspace{-1em} 
\begin{table}[htbp]
  \centering
    \caption{Comparison with other methods.}
  \label{tab:comparison}
  % \begin{adjustbox}{width=\columnwidth,center}
  % \small
  \begin{tabular}{l|ccc}
    \toprule
    \midrule
    Method & Sensor & Map & PS \\
    \midrule
    EKF \cite{ekf} & IMU + WSS & $\times$ & $\times$ \\
    A-LOAM \cite{aloam} & LiDAR & Point Cloud & $\times$ \\
    VISSLAM \cite{visslam} & Camera + IMU & Semantic & $\checkmark$ \\
    VINS \cite{vinsmono} & Camera + IMU + WSS & Geometric & $\times$ \\ \midrule
    Ours & Camera + IMU + WSS & Semantic & $\checkmark$ \\
      % Local Training Batchsize & 64 & 64 &  128\\
    \bottomrule
  \end{tabular}
  % \end{adjustbox}
\end{table}
% \vspace{-1em}

\textbf{PS Detector} 
Our PS detector is a YOLOv5 based network, which is specifically trained with nearly 50k BEV images collected in underground and outdoor parking lots for better generalization.  
Our PSDet module views each parking slot (PS) as a planar object with corresponding human-annotated label (i.e. PS shape, PS occupation state, 4 corner point location and related visible states), see Fig.\ref{fig:methodd}, \ref{fig:opt} and \ref{fig:rviz}.
% Each PS ground truth in BEV image is human-annotated with its visible corner points (or marking points \cite{maskRCNN,DeepPS}) and the PS status (i.e. vacant in green or occupied in red, see Fig.\ref{fig:methodd} and \ref{fig:opt}).
We only keep PS detection results with output probabilities greater than 50\%.
The training process lasts for 28 epochs with a fixed batchsize of 64. The well trained model is then deployed to support online inference via TensorRT \cite{tensorrt}.

\textbf{Evaluation Trajectories}
% In this work, we mainly focus on the underground autonomous parking scenarios. 
% Since there are no existing underground parking datasets which simultaneously contains IMU, WSS, LiDAR and surround-view camera these multi-modal sensor measurements to support evaluating diverse SLAM methods, 
We collect a comprehensive group of short-distance trajectories (APA scenarios) and another group of long-distance trajectories (AVP scenarios) in two real-world parking lots (see Tab.\ref{tab:main} and Fig.\ref{fig:traj}, $\#1$ denotes the first one and $\#2$ denotes another). The vehicle moves at the speed of about 4 km/h for short-distance trajectories, and about 10 km/h for long-distance ones.  During each trajectory, we record all sensors' perception measurements into ROS bags \cite{ROS} for offline evaluation. We prepare Tab.\ref{tab:lenNsize} to show lengths and duration time of trajectories.
\vspace{-1em}
\vspace{-1em}
\vspace{-1em}  
\begin{table}[htbp]
\caption{Lengths (with our errors) and duration of all trajectories.}
\label{tab:lenNsize}
\begin{adjustbox}{width=\columnwidth,center}
\begin{tabular}{cc|c|c|c|c|c|c|c|c}
\toprule
\midrule
 &  & straight 0$^\circ$ & left 45$^\circ$ & right 45$^\circ$ & left 90$^\circ$ & right 90$^\circ$ & left parallel & right parallel & round \\ \hline
\multicolumn{1}{c|}{\multirow{2}{*}{\begin{tabular}[c]{@{}c@{}}Traj. length with\\  our errors (meter)\end{tabular}}} & \quad Parking lot\#1 & 4.94 (0.13) & 5.95 (0.13) & 10.92 (0.11) & 11.29 (0.23) & 10.24 (0.06) & 8.25 (0.05) & 13.08 (0.18) & 121.62 (1.09) \\ \cline{2-10} 
\multicolumn{1}{c|}{} & \quad Parking lot\#2 & 4.64 (0.05) & 5.81 (0.01) & 6.00 (0.07) & 8.33 (0.29) & 12.35 (0.21) & 6.53 (0.07) & 7.91 (0.17) & 133.50 (1.27) \\ \hline
\multicolumn{1}{c|}{\multirow{2}{*}{\begin{tabular}[c]{@{}c@{}}Duration \\ (second)\end{tabular}}} & \quad Parking lot\#1 & 24.1 & 28.7 & 52.3 & 49.1 & 52.2 & 69.1 & 74.5 & 92.0 \\ \cline{2-10} 
\multicolumn{1}{c|}{} & \quad Parking lot\#2 & 24.9 & 28.8 & 39.7 & 36.3 & 56.5 & 31.5 & 38.2 & 102.2 \\ 
% \multicolumn{1}{c|}{\multirow{2}{*}{\begin{tabular}[c]{@{}c@{}}Data size\\ (GB)\end{tabular}}} & Parking lot\#1 & 12.7 & 15.1 & 27.5 & 25.8 & 27.4 & 36.3 & 39.2 & 48.3 \\ \cline{2-10} 
% \multicolumn{1}{c|}{} & Parking lot\#2 & 13.1 & 15.2 & 20.9 & 19.1 & 29.7 & 16.6 & 20.1 & 53.8 \\ 
\midrule 
\bottomrule
\end{tabular}
\end{adjustbox}
\end{table}
\vspace{-1em} 

\textbf{Ground Truth Acquisition \& Metric}
We use Root Mean Square Error (RMSE) which is a commonly used error metric for measuring trajectory errors. Most methods use GPS/RTK signal as ground truth in the open outdoor area, but it is not suitable for underground environments which have poor coverage of GPS/RTK measurements. Therefore, we refer to utilize motion capture system to acquire trajectories' ground truth by following previous works \cite{tum}.

\subsection{Experimental Results}
We compare our method with other three baseline methods for various parking trajectories, including short-distance parking scenarios and long-distance parking scenarios. Experimental results are shown in Tab.\ref{tab:main}. The corresponding parking trajectories are shown in Fig.\ref{fig:traj}.

For short-distance scenarios, our VIPS-Odom achieves better overall localization performance. Merging parking slot features to the visual-inertial odometry can effectively reduce localization error. Meanwhile, A-LOAM has comparable results with ours though LiDAR is much costly than our sensor configuration, which reflects the potential of using optimization-based methods with low-cost sensors. For long-distance scenarios, optimization-based methods, VISSLAM, VINS and VIPS-Odom, achieve better localization performance. VISSLAM got a failure case on one long-distance trajectory and we analyze the reason in Sec.\ref{sec:visslam}. With semantic constraints provided by PS, our VIPS-Odom has lower localization error than VINS, which indicates the drifts are efficiently reduced by incorporating PS.
\begin{table}[t]
  \centering
    \caption{Localization errors ($\downarrow$, unit: meter) on multiple trajectories. \textbf{The bold} denotes the best localization performance and \underline{the underlined} denotes the second performance.}
  \label{tab:main}
  \begin{adjustbox}{width=\columnwidth,center}
% \begin{table}[]
\begin{tabular}{ccccccc}
\hline \hline
\multicolumn{2}{c}{} & \multicolumn{5}{c}{\textbf{Method}} \\ \cline{3-7} 
\multicolumn{2}{c}{\multirow{-2}{*}{\textbf{Trajectory}}} & EKF & LOAM & \multicolumn{1}{l}{VISSLAM} & VINS & Ours \\ \hline
\multicolumn{1}{c|}{} & \multicolumn{1}{c|}{straight 0$^\circ$ (\#1)} & 0.14 & 0.48 & 0.39 & \textbf{0.11} & {\ul 0.13} \\
\multicolumn{1}{c|}{} & \multicolumn{1}{c|}{straight 0$^\circ$ (\#2)} & 0.11 & 0.12 & {\ul 0.08} & {\ul 0.08} & \textbf{0.05} \\ \cline{2-7} 
\multicolumn{1}{c|}{} & \multicolumn{1}{c|}{left 45$^\circ$ (\#1)} & 0.17 & \textbf{0.11} & 0.20 & 0.16 & {\ul 0.13} \\
\multicolumn{1}{c|}{} & \multicolumn{1}{c|}{left 45$^\circ$ (\#2)} & 0.13 & 0.16 & {\ul 0.03} & 0.16 & \textbf{0.01} \\ \cline{2-7} 
\multicolumn{1}{c|}{} & \multicolumn{1}{c|}{right 45$^\circ$ (\#1)} & 0.12 & \textbf{0.06} & 0.12 & 0.18 & {\ul 0.11} \\
\multicolumn{1}{c|}{} & \multicolumn{1}{c|}{right 45$^\circ$ (\#2)} & 0.16 & 0.12 & \textbf{0.07} & \textbf{0.07} & \textbf{0.07} \\ \cline{2-7} 
\multicolumn{1}{c|}{} & \multicolumn{1}{c|}{left 90$^\circ$ (\#1)} & 0.26 & \textbf{0.04} & 0.27 & {\ul 0.18} & 0.23 \\
\multicolumn{1}{c|}{} & \multicolumn{1}{c|}{left 90$^\circ$ (\#2)} & {\ul 0.25} & \textbf{0.18} & 0.36 & 0.35 & 0.29 \\ \cline{2-7} 
\multicolumn{1}{c|}{} & \multicolumn{1}{c|}{right 90$^\circ$ (\#1)} & 0.10 & 0.13 & 0.09 & {\ul 0.07} & \textbf{0.06} \\
\multicolumn{1}{c|}{} & \multicolumn{1}{c|}{right 90$^\circ$ (\#2)} & {\ul 0.11} & \textbf{0.07} & 0.23 & 0.23 & 0.21 \\ \cline{2-7} 
\multicolumn{1}{c|}{} & \multicolumn{1}{c|}{left parallel (\#1)} & 0.11 & {\ul 0.05} & \textbf{0.03} & 0.10 & {\ul 0.05} \\
\multicolumn{1}{c|}{} & \multicolumn{1}{c|}{left parallel (\#2)} & 0.17 & 0.11 & 0.13 & {\ul 0.08} & \textbf{0.07} \\ \cline{2-7} 
\multicolumn{1}{c|}{} & \multicolumn{1}{c|}{right parallel (\#1)} & {\ul 0.16} & \textbf{0.04} & 0.22 & 0.19 & 0.18 \\
\multicolumn{1}{c|}{\multirow{-14}{*}{\textbf{Short-distance}}} & \multicolumn{1}{c|}{right parallel (\#2)} & 0.17 & \textbf{0.14} & 0.23 & {\ul 0.16} & 0.17 \\ \cline{2-7} 
\multicolumn{1}{c|}{} & \multicolumn{1}{c|}{\cellcolor[HTML]{EFEFEF}Overall mean} & \cellcolor[HTML]{EFEFEF}0.154 & \cellcolor[HTML]{EFEFEF}{\ul 0.129} & \cellcolor[HTML]{EFEFEF}0.175 & \cellcolor[HTML]{EFEFEF}0.151 & \cellcolor[HTML]{EFEFEF}\textbf{0.126} \\ \hline
\multicolumn{1}{c|}{} & \multicolumn{1}{c|}{round (\#1)} & 0.35 & 7.60 & fail & 2.21 & 1.09 \\
\multicolumn{1}{c|}{} & \multicolumn{1}{c|}{round (\#2)} & 7.86 & 2.73 & 1.73 & 1.85 & 1.27 \\ \cline{2-7} 
\multicolumn{1}{c|}{\multirow{-3}{*}{\textbf{Long-distance}}} & \multicolumn{1}{c|}{\cellcolor[HTML]{EFEFEF}Overall mean} & \cellcolor[HTML]{EFEFEF}4.105 & \cellcolor[HTML]{EFEFEF}5.165 & \cellcolor[HTML]{EFEFEF}- & \cellcolor[HTML]{EFEFEF}{\ul 2.030} & \cellcolor[HTML]{EFEFEF}\textbf{1.180} \\ \hline
\multicolumn{1}{c|}{\textbf{Oveall}} & \multicolumn{1}{c|}{\cellcolor[HTML]{C0C0C0}Overall mean} & \cellcolor[HTML]{C0C0C0}0.648 & \cellcolor[HTML]{C0C0C0}0.759 & \cellcolor[HTML]{C0C0C0}- & \cellcolor[HTML]{C0C0C0}{\ul 0.386} & \cellcolor[HTML]{C0C0C0}\textbf{0.258} \\ \hline
\end{tabular}
% \end{table}
  \end{adjustbox}

\end{table}

\begin{table}[t]
  \centering
    \caption{Ablation results (unit: Meter) on only keeping parking slots information in frontend (front.) or backend (back.).}
  \label{tab:comparison_abla}
  \begin{adjustbox}{width=\columnwidth,center}
\begin{tabular}{cccccc}
% \hline
\toprule
\midrule
\multicolumn{2}{c}{}                                                                                             & \multicolumn{4}{c}{\textbf{Method}}                                                                                                                                                                      \\ \cline{3-6} 
\multicolumn{2}{c}{\multirow{-2}{*}{\textbf{Trajectory}}}                                                        & VINS                          & \begin{tabular}[c]{@{}c@{}}Ours\\ (+front.)\end{tabular} & \begin{tabular}[c]{@{}c@{}}Ours\\ (+back.)\end{tabular} & \begin{tabular}[c]{@{}c@{}}Ours\\ (+both)\end{tabular} \\ \hline
\multicolumn{1}{c|}{}                                & \multicolumn{1}{c|}{Straight 0$^\circ$ (\#1)}             & 0.11                          & \textbf{0.03}                                           & 0.12                                                   & 0.13                                                  \\
\multicolumn{1}{c|}{}                                & \multicolumn{1}{c|}{Straight 0$^\circ$ (\#2)}             & 0.08                          & 0.07                                                    & 0.07                                                   & \textbf{0.05}                                         \\ \cline{2-6} 
\multicolumn{1}{c|}{}                                & \multicolumn{1}{c|}{Left 45$^\circ$ (\#1)}                & 0.15                          & 0.15                                                    & 0.14                                                   & 0.13                                                  \\
\multicolumn{1}{c|}{}                                & \multicolumn{1}{c|}{Left 45$^\circ$ (\#2)}                & 0.04                          & 0.02                                                    & 0.04                                                   & \textbf{0.01}                                         \\ \cline{2-6} 
\multicolumn{1}{c|}{}                                & \multicolumn{1}{c|}{Right 45$^\circ$ (\#1)}               & 0.18                          & 0.17                                                    & \textbf{0.13}                                          & 0.11                                                  \\
\multicolumn{1}{c|}{}                                & \multicolumn{1}{c|}{Right 45$^\circ$ (\#2)}               & 0.07                          & 0.07                                                    & \textbf{0.06}                                          & 0.07                                                  \\ \cline{2-6} 
\multicolumn{1}{c|}{}                                & \multicolumn{1}{c|}{Left 90$^\circ$ (\#1)}                & \textbf{0.18}                 & 0.26                                                    & 0.31                                                   & 0.23                                                  \\
\multicolumn{1}{c|}{}                                & \multicolumn{1}{c|}{Left 90$^\circ$ (\#2)}                & 0.35                          & 0.30                                                    & 0.32                                                   & \textbf{0.29}                                         \\ \cline{2-6} 
\multicolumn{1}{c|}{}                                & \multicolumn{1}{c|}{Right 90$^\circ$ (\#1)}               & 0.07                          & \textbf{0.03}                                           & 0.11                                                   & 0.06                                                  \\
\multicolumn{1}{c|}{}                                & \multicolumn{1}{c|}{Right 90$^\circ$ (\#2)}               & 0.23                          & 0.20                                                    & \textbf{0.12}                                          & 0.21                                                  \\ \cline{2-6} 
\multicolumn{1}{c|}{}                                & \multicolumn{1}{c|}{Left parallel (\#1)}                  & 0.10                          & \textbf{0.04}                                           & 0.09                                                   & 0.05                                                  \\
\multicolumn{1}{c|}{}                                & \multicolumn{1}{c|}{Left parallel (\#2)}                  & 0.08                          & 0.07                                                    & \textbf{0.06}                                          & 0.07                                                  \\ \cline{2-6} 
\multicolumn{1}{c|}{}                                & \multicolumn{1}{c|}{Right parallel (\#1)}                 & 0.19                          & 0.20                                                    & 0.20                                                   & \textbf{0.18}                                         \\
\multicolumn{1}{c|}{\multirow{-14}{*}{\textbf{Short-distance}}} & \multicolumn{1}{c|}{Right parallel (\#2)}                 & 0.16                          & \textbf{0.14}                                           & 0.19                                                   & 0.17                                                  \\ \cline{2-6} 
\multicolumn{1}{c|}{}                                & \multicolumn{1}{c|}{\cellcolor[HTML]{EFEFEF}Mean} & \cellcolor[HTML]{EFEFEF}0.142 & \cellcolor[HTML]{EFEFEF}\textbf{0.125}                  & \cellcolor[HTML]{EFEFEF}0.140                          & \cellcolor[HTML]{EFEFEF}0.126                         \\ \hline
\multicolumn{1}{c|}{}                                & \multicolumn{1}{c|}{Round (\#1)}                          & 2.21                          & 2.22                                                    & 1.60                                                   & \textbf{1.09}                                         \\
\multicolumn{1}{c|}{}                                & \multicolumn{1}{c|}{Round (\#2)}                          & 1.85                          & 1.81                                                    & \textbf{1.26}                                          & 1.27                                                  \\ \cline{2-6} 
\multicolumn{1}{c|}{\multirow{-3}{*}{\textbf{Long-distance}}}  & \multicolumn{1}{c|}{\cellcolor[HTML]{EFEFEF}Mean} & \cellcolor[HTML]{EFEFEF}2.03  & \cellcolor[HTML]{EFEFEF}2.02                            & \cellcolor[HTML]{EFEFEF}1.43                           & \cellcolor[HTML]{EFEFEF}\textbf{1.18}                 \\ \hline
\multicolumn{1}{c|}{\textbf{Overall}}                & \multicolumn{1}{c|}{\cellcolor[HTML]{C0C0C0}Mean} & \cellcolor[HTML]{C0C0C0}0.378 & \cellcolor[HTML]{C0C0C0}0.361                           & \cellcolor[HTML]{C0C0C0}0.301                          & \cellcolor[HTML]{C0C0C0}\textbf{0.258}                \\ \hline
\end{tabular}
  \end{adjustbox}
\end{table}
\vspace{-1em}
\begin{table*}[htbp]
  \centering
    \caption{Experimental results (unit: Meter) on sensitivity analysis. Std. denotes the standard deviation among each group.}
  \label{tab:comparison_sens}
  \begin{adjustbox}{width=\textwidth,center}
  
\begin{tabular}{cccccccccc|cccccccc}
\toprule
\midrule
\multicolumn{2}{c}{}                                                                                             & \multicolumn{8}{c|}{\textbf{Feature Point Number \#N (Fixed K=10)}}                                                                                                                                                                                                                                                                            & \multicolumn{8}{c}{\textbf{Keyframes Number \#K (Fixed N=110)}}                                                                                                                                                                                                                                                                       \\ \cline{3-18} 
\multicolumn{2}{c}{}                                                                                             & \multicolumn{4}{c|}{VINS}                                                                                                                                                        & \multicolumn{4}{c|}{Ours}                                                                                                                                   & \multicolumn{4}{c|}{VINS}                                                                                                                                               & \multicolumn{4}{c}{Ours}                                                                                                                                    \\ \cline{3-18} 
\multicolumn{2}{c}{\multirow{-3}{*}{\textbf{Trajectory}}}                                                        & N=50                          & N=110                         & \multicolumn{1}{c|}{N=200}                         & \multicolumn{1}{c|}{std.}                                   & N=50                          & N=110                         & \multicolumn{1}{c|}{N=200}                         & std.                                   & K=7                           & K=10                          & \multicolumn{1}{c|}{K=13}                          & \multicolumn{1}{c|}{std.}                          & K=7                           & K=10                          & \multicolumn{1}{c|}{K=13}                          & std.                                   \\ \hline
\multicolumn{1}{c|}{}                                & \multicolumn{1}{c|}{Straight 0$^\circ$ (\#1)}             & 0.08                          & 0.11                          & \multicolumn{1}{c|}{0.11}                          & \multicolumn{1}{c|}{\textbf{0.02}}                          & 0.08                          & 0.13                          & \multicolumn{1}{c|}{0.12}                          & 0.03                                   & 0.11                          & 0.11                          & \multicolumn{1}{c|}{0.14}                          & \multicolumn{1}{c|}{0.02}                          & 0.11                          & 0.13                          & \multicolumn{1}{c|}{0.11}                          & \textbf{0.01}                          \\
\multicolumn{1}{c|}{}                                & \multicolumn{1}{c|}{Straight 0$^\circ$ (\#2)}             & 0.05                          & 0.08                          & \multicolumn{1}{c|}{0.08}                          & \multicolumn{1}{c|}{0.02}                                   & 0.05                          & 0.05                          & \multicolumn{1}{c|}{0.06}                          & \textbf{0.01}                          & 0.08                          & 0.08                          & \multicolumn{1}{c|}{0.07}                          & \multicolumn{1}{c|}{\textbf{0.01}}                 & 0.07                          & 0.05                          & \multicolumn{1}{c|}{0.07}                          & \textbf{0.01}                          \\ \cline{2-18} 
\multicolumn{1}{c|}{}                                & \multicolumn{1}{c|}{Left 45$^\circ$ (\#1)}                & 0.13                          & 0.15                          & \multicolumn{1}{c|}{0.18}                          & \multicolumn{1}{c|}{0.03}                                   & 0.14                          & 0.13                          & \multicolumn{1}{c|}{0.13}                          & \textbf{0.01}                          & 0.20                          & 0.15                          & \multicolumn{1}{c|}{0.12}                          & \multicolumn{1}{c|}{0.04}                          & 0.13                          & 0.13                          & \multicolumn{1}{c|}{0.12}                          & \textbf{0.01}                          \\
\multicolumn{1}{c|}{}                                & \multicolumn{1}{c|}{Left 45$^\circ$ (\#2)}                & 0.04                          & 0.04                          & \multicolumn{1}{c|}{0.10}                          & \multicolumn{1}{c|}{\textbf{0.04}}                          & 0.02                          & 0.01                          & \multicolumn{1}{c|}{0.09}                          & \textbf{0.04}                          & 0.07                          & 0.04                          & \multicolumn{1}{c|}{0.04}                          & \multicolumn{1}{c|}{\textbf{0.02}}                 & 0.04                          & 0.01                          & \multicolumn{1}{c|}{0.01}                          & \textbf{0.02}                          \\ \cline{2-18} 
\multicolumn{1}{c|}{}                                & \multicolumn{1}{c|}{Right 45$^\circ$ (\#1)}               & 0.11                          & 0.18                          & \multicolumn{1}{c|}{0.10}                          & \multicolumn{1}{c|}{0.04}                                   & 0.12                          & 0.11                          & \multicolumn{1}{c|}{0.14}                          & \textbf{0.02}                          & 0.11                          & 0.18                          & \multicolumn{1}{c|}{0.27}                          & \multicolumn{1}{c|}{0.08}                          & 0.17                          & 0.11                          & \multicolumn{1}{c|}{0.14}                          & \textbf{0.03}                          \\
\multicolumn{1}{c|}{}                                & \multicolumn{1}{c|}{Right 45$^\circ$ (\#2)}               & 0.07                          & 0.07                          & \multicolumn{1}{c|}{0.08}                          & \multicolumn{1}{c|}{\textbf{0.01}}                          & 0.06                          & 0.07                          & \multicolumn{1}{c|}{0.06}                          & \textbf{0.01}                          & 0.07                          & 0.07                          & \multicolumn{1}{c|}{0.08}                          & \multicolumn{1}{c|}{\textbf{0.01}}                 & 0.06                          & 0.07                          & \multicolumn{1}{c|}{0.07}                          & \textbf{0.01}                          \\ \cline{2-18} 
\multicolumn{1}{c|}{}                                & \multicolumn{1}{c|}{Left 90$^\circ$ (\#1)}                & 0.14                          & 0.18                          & \multicolumn{1}{c|}{0.19}                          & \multicolumn{1}{c|}{\textbf{0.03}}                          & 0.20                          & 0.23                          & \multicolumn{1}{c|}{0.28}                          & 0.04                                   & 0.18                          & 0.18                          & \multicolumn{1}{c|}{0.15}                          & \multicolumn{1}{c|}{0.02}                          & 0.24                          & 0.23                          & \multicolumn{1}{c|}{0.24}                          & \textbf{0.01}                          \\
\multicolumn{1}{c|}{}                                & \multicolumn{1}{c|}{Left 90$^\circ$ (\#2)}                & 0.28                          & 0.35                          & \multicolumn{1}{c|}{0.33}                          & \multicolumn{1}{c|}{0.04}                                   & 0.32                          & 0.29                          & \multicolumn{1}{c|}{0.34}                          & \textbf{0.03}                          & 0.31                          & 0.35                          & \multicolumn{1}{c|}{0.32}                          & \multicolumn{1}{c|}{0.02}                          & 0.30                          & 0.29                          & \multicolumn{1}{c|}{0.30}                          & \textbf{0.01}                          \\ \cline{2-18} 
\multicolumn{1}{c|}{}                                & \multicolumn{1}{c|}{Right 90$^\circ$ (\#1)}               & 0.04                          & 0.07                          & \multicolumn{1}{c|}{0.07}                          & \multicolumn{1}{c|}{\textbf{0.02}}                          & 0.09                         & 0.06                          & \multicolumn{1}{c|}{0.08}                          & \textbf{0.02}                                   & 0.05                          & 0.07                          & \multicolumn{1}{c|}{0.07}                          & \multicolumn{1}{c|}{\textbf{0.01}}                 & 0.15                          & 0.06                          & \multicolumn{1}{c|}{0.12}                          & 0.05                                   \\
\multicolumn{1}{c|}{}                                & \multicolumn{1}{c|}{Right 90$^\circ$ (\#2)}               & 0.23                          & 0.23                          & \multicolumn{1}{c|}{0.22}                          & \multicolumn{1}{c|}{\textbf{0.01}}                          & 0.28                          & 0.21                          & \multicolumn{1}{c|}{0.16}                          & 0.06                                   & 0.19                          & 0.23                          & \multicolumn{1}{c|}{0.24}                          & \multicolumn{1}{c|}{0.03}                          & 0.21                          & 0.21                          & \multicolumn{1}{c|}{0.23}                          & \textbf{0.01}                          \\ \cline{2-18} 
\multicolumn{1}{c|}{}                                & \multicolumn{1}{c|}{Left parallel (\#1)}                  & 0.10                          & 0.10                          & \multicolumn{1}{c|}{0.14}                          & \multicolumn{1}{c|}{\textbf{0.02}}                          & 0.04                          & 0.05                          & \multicolumn{1}{c|}{0.01}                          & \textbf{0.02}                          & 0.43                          & 0.10                          & \multicolumn{1}{c|}{0.12}                          & \multicolumn{1}{c|}{0.19}                          & 0.01                          & 0.05                          & \multicolumn{1}{c|}{0.01}                          & \textbf{0.02}                          \\
\multicolumn{1}{c|}{}                                & \multicolumn{1}{c|}{Left parallel (\#2)}                  & 0.08                          & 0.08                          & \multicolumn{1}{c|}{0.07}                          & \multicolumn{1}{c|}{\textbf{0.01}}                          & 0.01                          & 0.07                          & \multicolumn{1}{c|}{0.05}                          & 0.03                                   & 0.10                          & 0.08                          & \multicolumn{1}{c|}{0.10}                          & \multicolumn{1}{c|}{\textbf{0.01}}                 & 0.08                          & 0.07                          & \multicolumn{1}{c|}{0.10}                          & 0.02                                   \\ \cline{2-18} 
\multicolumn{1}{c|}{}                                & \multicolumn{1}{c|}{Right parallel (\#1)}                 & 0.18                          & 0.19                          & \multicolumn{1}{c|}{0.15}                          & \multicolumn{1}{c|}{\textbf{0.02}}                          & 0.16                          & 0.18                          & \multicolumn{1}{c|}{0.24}                          & 0.04                                   & 0.15                          & 0.19                          & \multicolumn{1}{c|}{0.15}                          & \multicolumn{1}{c|}{0.02}                          & 0.20                          & 0.18                          & \multicolumn{1}{c|}{0.20}                          & \textbf{0.01}                          \\
\multicolumn{1}{c|}{\multirow{-14}{*}{\textbf{Short-distance}}} & \multicolumn{1}{c|}{Right parallel (\#2)}                 & 0.07                          & 0.16                          & \multicolumn{1}{c|}{0.15}                          & \multicolumn{1}{c|}{0.05}                                   & 0.13                          & 0.17                          & \multicolumn{1}{c|}{0.15}                          & \textbf{0.02}                          & 0.14                          & 0.16                          & \multicolumn{1}{c|}{0.11}                          & \multicolumn{1}{c|}{0.03}                          & 0.16                          & 0.17                          & \multicolumn{1}{c|}{0.15}                          & \textbf{0.01}                          \\ \cline{2-18} 
\multicolumn{1}{c|}{}                                & \multicolumn{1}{c|}{\cellcolor[HTML]{EFEFEF}Mean} & \cellcolor[HTML]{EFEFEF}0.114 & \cellcolor[HTML]{EFEFEF}0.142 & \multicolumn{1}{c|}{\cellcolor[HTML]{EFEFEF}0.141} & \multicolumn{1}{c|}{\cellcolor[HTML]{EFEFEF}\textbf{0.026}} & \cellcolor[HTML]{EFEFEF}0.121 & \cellcolor[HTML]{EFEFEF}0.126 & \multicolumn{1}{c|}{\cellcolor[HTML]{EFEFEF}0.136} & \cellcolor[HTML]{EFEFEF}0.027          & \cellcolor[HTML]{EFEFEF}0.156 & \cellcolor[HTML]{EFEFEF}0.142 & \multicolumn{1}{c|}{\cellcolor[HTML]{EFEFEF}0.141} & \multicolumn{1}{c|}{\cellcolor[HTML]{EFEFEF}0.036} & \cellcolor[HTML]{EFEFEF}0.138 & \cellcolor[HTML]{EFEFEF}0.126 & \multicolumn{1}{c|}{\cellcolor[HTML]{EFEFEF}0.134} & \cellcolor[HTML]{EFEFEF}\textbf{0.016} \\ \hline
\multicolumn{1}{c|}{}                                & \multicolumn{1}{c|}{Round (\#1)}                          & 2.27                          & 2.21                          & \multicolumn{1}{c|}{2.64}                          & \multicolumn{1}{c|}{0.23}                                   & 1.33                          & 1.09                          & \multicolumn{1}{c|}{1.25}                          & \textbf{0.12}                          & 2.53                          & 2.21                          & \multicolumn{1}{c|}{2.08}                          & \multicolumn{1}{c|}{0.23}                          & 0.87                          & 1.09                          & \multicolumn{1}{c|}{1.01}                          & \textbf{0.11}                          \\
\multicolumn{1}{c|}{}                                & \multicolumn{1}{c|}{Round (\#2)}                          & 1.93                          & 1.85                          & \multicolumn{1}{c|}{1.65}                          & \multicolumn{1}{c|}{0.14}                                   & 1.29                          & 1.27                          & \multicolumn{1}{c|}{1.33}                          & \textbf{0.03}                          & 1.74                          & 1.85                          & \multicolumn{1}{c|}{1.78}                          & \multicolumn{1}{c|}{\textbf{0.06}}                 & 1.09                          & 1.27                          & \multicolumn{1}{c|}{1.30}                          & 0.11                                   \\ \cline{2-18} 
\multicolumn{1}{c|}{\multirow{-3}{*}{\textbf{Long-distance}}}  & \multicolumn{1}{c|}{\cellcolor[HTML]{EFEFEF}Mean} & \cellcolor[HTML]{EFEFEF}2.10  & \cellcolor[HTML]{EFEFEF}2.03  & \multicolumn{1}{c|}{\cellcolor[HTML]{EFEFEF}2.145} & \multicolumn{1}{c|}{\cellcolor[HTML]{EFEFEF}0.185}          & \cellcolor[HTML]{EFEFEF}1.31  & \cellcolor[HTML]{EFEFEF}1.18  & \multicolumn{1}{c|}{\cellcolor[HTML]{EFEFEF}1.29}  & \cellcolor[HTML]{EFEFEF}\textbf{0.075} & \cellcolor[HTML]{EFEFEF}2.135 & \cellcolor[HTML]{EFEFEF}2.03  & \multicolumn{1}{c|}{\cellcolor[HTML]{EFEFEF}1.93}  & \multicolumn{1}{c|}{\cellcolor[HTML]{EFEFEF}0.145} & \cellcolor[HTML]{EFEFEF}0.98  & \cellcolor[HTML]{EFEFEF}1.18  & \multicolumn{1}{c|}{\cellcolor[HTML]{EFEFEF}1.155} & \cellcolor[HTML]{EFEFEF}\textbf{0.110} \\ \hline
\multicolumn{1}{c|}{\textbf{Overall}}                & \multicolumn{1}{c|}{\cellcolor[HTML]{C0C0C0}Mean} & \cellcolor[HTML]{C0C0C0}0.363 & \cellcolor[HTML]{C0C0C0}0.378 & \multicolumn{1}{c|}{\cellcolor[HTML]{C0C0C0}0.391} & \multicolumn{1}{c|}{\cellcolor[HTML]{C0C0C0}0.046}          & \cellcolor[HTML]{C0C0C0}0.270 & \cellcolor[HTML]{C0C0C0}0.258 & \multicolumn{1}{c|}{\cellcolor[HTML]{C0C0C0}0.281} & \cellcolor[HTML]{C0C0C0}\textbf{0.033} & \cellcolor[HTML]{C0C0C0}0.404 & \cellcolor[HTML]{C0C0C0}0.378 & \multicolumn{1}{c|}{\cellcolor[HTML]{C0C0C0}0.365} & \multicolumn{1}{c|}{\cellcolor[HTML]{C0C0C0}0.050} & \cellcolor[HTML]{C0C0C0}0.243 & \cellcolor[HTML]{C0C0C0}0.258 & \multicolumn{1}{c|}{\cellcolor[HTML]{C0C0C0}0.261} & \cellcolor[HTML]{C0C0C0}\textbf{0.028} \\ \hline
\end{tabular}

  \end{adjustbox}

\end{table*}

\subsection{Ablation Study}
In this work, we fuse the parking slot (PS) information into both SLAM frontend and backend, and herein we conduct ablation study to investigate the role of PS information in the frontend or backend. The experimental results are shown in Tab.\ref{tab:comparison_abla}. Based on these results, we find the incorporating of PS in the frontend benefits the short-distance localization during short-distance parking trajectories, and the incorporating of PS in the backend benefits the reduction of long-distance accumulated drift over long-distance trajectories and improve the localization precision. 
In short-distance (near the target parking slot) parking scenarios, the parking slot features supplemented by the frontend improve the overall quality of the feature points, helping the algorithm to better track feature points and estimate pose in the backend. In long-distance scenarios, vehicles will cruise at a relatively steady speed. At this time, it is particularly important to add parking slot as extra semantic objects in the backend optimization to reduce long-term accumulated drifts.

% \vspace{-1em}
% \vspace{-1em}

\subsection{Sensitivity Analysis}
\label{sec:comparison_robust}
When deployed in practice, optimization-based SLAM methods have two important hyper-parameters including keyframe number of each sliding window and feature point number of tracked images. Herein we conduct sensitivity analysis on these two hyper-parameters to demonstrate the robustness (or sensitivity) of our VIPS-Odom against another optimization-based method VINS \cite{vinsmono}. Related results in shown in Tab.\ref{tab:comparison_sens}. 
For feature point number selection, we choose different number of maximum feature points in the front-view camera to investigate its impact on the localization accuracy. With the assistance of PS, our VIPS-Odom is more robust to the selection of feature point in long-distance trajectories. We also observe that VIPS-Odom is slightly less robust than VINS in short-distance trajectories. Meanwhile, the selection of keyframe number in the optimization sliding window is of significance in the optimization-based methods.
We choose different number of keyframes in the backend optimization to investigate its impact on the localization accuracy. As the results shown, our VIPS-Odom achieves better robustness on keyframe number selection. 
This helps VIPS-Odom have better generalization for more scenarios.
\vspace{-1em}  
\begin{figure}[htbp]
    \centering
    \includegraphics[width=\linewidth]{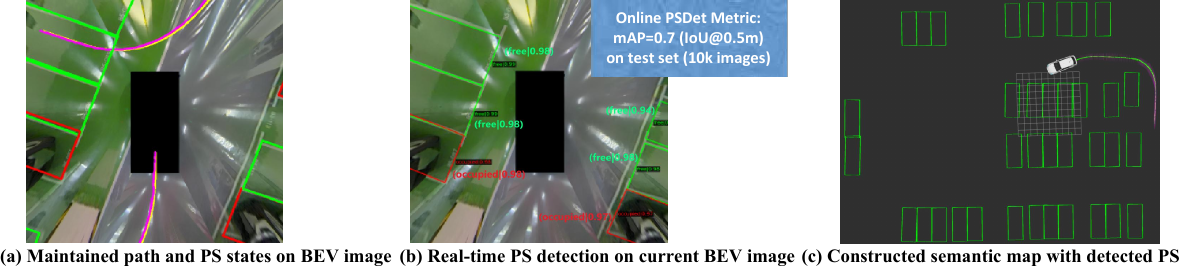}
    \caption{A brief qualitative illustration of our system.}
    \label{fig:rviz}
\end{figure}
% \vspace{-0.9em} 

\subsection{Qualitative Evaluation and Real-time Analysis}
With Rviz toolkit, we visualize one long-distance trajectory (\#1) and provide qualitative evaluation of our VIPS-Odom in Fig.\ref{fig:rviz}. Our real-time PS detection module detects parking slots on the current frame. Furthermore, our PS management module tracks  all observed parking slots under a multi-object tracking framework. With the tracking framework, we robustly track all PS states to handle real-time PS observation and manage all PS states, thus maintained well-tracked PS information (Fig.\ref{fig:rviz}(a)) is considered in backend. Regarding frequency, the frequency of IMU/WSS readings are 100Hz, and the frequency of fisheye cameras is 20Hz. BEV image generation and PS Detection on generated BEV images are of 10Hz. The odometry is solved timely at 25Hz.
% \vspace{-1em}  
% \begin{figure}[htbp]
%     \centering
%     \includegraphics[width=\linewidth]{pics/qualitative results.pdf}
%     \caption{A brief qualitative illustration of our system.}
%     \label{fig:rviz}
% \end{figure}
% \vspace{-0.9em} 

% \begin{table}[hbtp]
%   \centering
%     \caption{Frequency Statistics.}
%   \label{tab:freq}
%   % \begin{adjustbox}{width=\columnwidth,center}
%   \small
%   \begin{tabular}{l|c|c|c|c}
%     \toprule
%     \midrule
%     IMU & WSS & Camera & BEV Image & PS Det. \\
%     \midrule
%     % EKF & IMU + WSS & $\times$ & $\times$ \\
%     100Hz & 100Hz & 20Hz & 10Hz & 10Hz \\ 
%       % Local Training Batchsize & 64 & 64 &  128\\
%     \bottomrule
%   \end{tabular}
%   % \end{adjustbox}
% \end{table}
% You must include your signed IEEE copyRight release form when you submit your finished paper.
% We MUST have this form before your paper can be published in the proceedings.

\subsection{PS Association Analysis}
\label{sec:visslam}
We compare the performance (RMSE errors) of our SORT based PS association method and hard match policy used in VISSLAM \cite{visslam}. Backend optimization is both performed by VIPS-Odom for a fair comparison. Our PS association is achieved by maintaining PS states in a SORT based real-time multi-object tracking framework. Hard match policy is achieved via strictly matching with PS observation center-point averaged value, which is prone to suffer long-distance noisy observation. We showcase experimental results in Tab.\ref{tab:comparison_PS}. For the long-distance Round (\#1) trajectory, hard match policy suffers from severe PS mismatches, creates numerous new parking slots in matching module, and encounters failure to form reliable PS factors (see \ref{sec:backend}). Above experiments demonstrate the flexibility and benefit of our PS association method against the hard match policy.
\vspace{-1em}
\begin{table}[htbp]
  \centering
    \caption{Results (unit: Meter) between two PS association methods. Thresholds \underline{th1} and \underline{th2} (see \cite{visslam}) for hard match are 0.5m/2m.}
  \label{tab:comparison_PS}
  % \begin{adjustbox}{width=\columnwidth,center}
  \small
  \begin{tabular}{l|cc}
    \toprule
    \midrule
    Trajectory & Hard Match Policy & SORT  \\
    \midrule
    % EKF & IMU + WSS & $\times$ & $\times$ \\
    Short-distance (Mean) & 0.165 & 0.126   \\ \midrule
    Long-distance Round (\#1) & fail & 1.09  \\ 
    Long-distance Round (\#2) & 1.49 & 1.27  \\
      % Local Training Batchsize & 64 & 64 &  128\\
    \bottomrule
  \end{tabular}
  % \end{adjustbox}
\end{table}
% \vspace{-1em} 

\subsection{Backend PS reweighting Analysis}
Herein we conduct ablation study on the utilization of reweighting PS error terms in Eq.\ref{eq:reg}. We modify the $\alpha_k^i$ in Eq.\ref{eq:reg} as constant $1$ to dismiss the reweighting terms. From experimental results, we find utilizing the reweighting terms can help to improve the localization accuracy.

\vspace{-1em}
\begin{table}[htbp]
  \centering
    \caption{Results (unit: Meter) on whether to utilize PS reweighting.}
  \label{tab:comparison_PS}
  \begin{adjustbox}{width=\columnwidth,center}
  \small
  \begin{tabular}{l|cc}
    \toprule
    \midrule
    Trajectory & without Reweighting & with Reweighting  \\
    \midrule
    % EKF & IMU + WSS & $\times$ & $\times$ \\
    Short-distance (Mean) & 0.138 & 0.126   \\ \midrule
    Long-distance Round (Mean) & 1.35 & 1.18  \\ 
    % Long-distance Round (\#2) & - & 1.27  \\
      % Local Training Batchsize & 64 & 64 &  128\\
    \bottomrule
  \end{tabular}
  \end{adjustbox}
\end{table}
% Please direct any questions to the production editor in charge of these proceedings at the IEEE Computer Society Press:
% \url{https://www.computer.org/about/contact}.

\section{Conclusion}
\label{sec:conc}
In this work, we propose a novel tightly-coupled SLAM system VIPS-Odom and fuse the measurements from an inertial measurement unit, a wheel speed sensor and four surround-view fisheye cameras to achieve high-precision localization with stability. We detect parking slots from BEV images in real time, robustly maintain observed parking slots' states, and simultaneously exploit parking slot observations as both frontend visual features and backend optimization factors to improve the localization precision. 
We also develop an experimental platform with the related sensor suite for evaluating the performance of VIPS-Odom.
Extensive real-world experiments covering both short-distance and long-distance parking scenarios reflect the effectiveness and stability of our method against other baseline methods. For future works, we consider to improve VIPS-Odom in the following aspects: 
(1) We do not include loop detection module into our system. Since loop detection is not necessary for all short-distance APA scenarios and partial long-distance AVP scenarios. We can extend loop detection module to complete our system to further improve the performance for long-distance scenarios.
(2) Beside the parking slots, We can incorporate more semantic objects into our system.

\end{document}